\documentclass[final]{IEEEtran}
\usepackage{times}

\usepackage[numbers]{natbib}
\usepackage{multicol}
\usepackage[bookmarks=true]{hyperref}
\usepackage{graphics}
\usepackage{amsmath}
\usepackage{amssymb}
\usepackage{algorithm}
\usepackage[noend]{algpseudocode}
\usepackage{booktabs}
\usepackage{color}
\usepackage{subcaption}
\usepackage{rotating}
\usepackage{xstring}
\usepackage{hhline}
\usepackage{siunitx}
\usepackage{physics}
\usepackage{tikz}
\usepackage{soul}
\usepackage{tcolorbox}
\usepackage{textcomp}
\usepackage{array,multirow,graphicx}
\usepackage{stmaryrd}

\begin{document}

\title{Outlier-Robust Long-Term Robotic Mapping Leveraging Ground Segmentation}

\author{Hyungtae Lim\\Massachusetts Institute of Technology}

\markboth{This research statement has been accepted for the RSS Pioneers 2024. The Pioneer lists can be viewed here: \href{https://sites.google.com/view/rsspioneers2024/participants}{https://sites.google.com/view/rsspioneers2024/participants}}
\maketitle


%
\def\secref#1{Section~\ref{#1}}
\def\figref#1{Fig.~\ref{#1}}
\def\tabref#1{Table~\ref{#1}}
\def\eqref#1{(\ref{#1})}

\maketitle

\IEEEpeerreviewmaketitle

\section{Motivation}

Despite the remarkable advancements in deep learning-based perception technologies~\cite{lee2022learning, lim2020msdpn, sung2024contextrast} and simultaneous localization and mapping~(SLAM)~\cite{cattaneo2022lcdnet, chen2022overlapnet, fischer2021stickypillars,komorowski2021minkloc3d, lee20232},
one can face the failure of these approaches when robots encounter scenarios outside their modeled experiences~\cite{lee2024object}~(here, the term \textit{modeling} encompasses both conventional pattern finding and data-driven approaches).
In particular, because learning-based methods are prone to catastrophic failure when operated in untrained scenes~\cite{lim2023quatro++,shaheen2022continual},
there is still a demand for conventional yet robust approaches that work out of the box in diverse scenarios~\cite{kim2018scancontext, lim2023adalio, qin2018vins, ramezani2022wildcat, segal2009gicp, song2022ral, vizzo2023kiss, wei2022fastlio}, such as real-world robotic services~\cite{lee2021assistive} and SLAM competitions~\cite{delmerico2019we,helmberger2022hilti,nair2024hilti, wang2020tartanair}.
In addition, the dynamic nature of real-world environments, characterized by changing surroundings over time and the presence of moving objects, leads to undesirable data points that hinder a robot from localization and path planning~\cite{kim2020iros}.
Consequently, methodologies that enable long-term map management~\cite{pomerleau2014icra}, such as multi-session SLAM~\cite{chang2022lamp, fernandez2024multi, tian2022kimera, zou2024lta} and static map building~\cite{jia2024beautymap, lim2021ral}, become essential.

Therefore, to achieve a robust long-term robotic mapping system that can work out of the box, first, I propose (i)~fast and robust ground segmentation to reject the ground points, which are featureless and thus not helpful for localization and mapping~\cite{lee2022patchworkpp, lim2021ral-patch}.
Then, by employing the concept of graduated non-convexity~(GNC)~\cite{yang2020gnc, yang2020teaser},
I propose (ii)~outlier-robust registration with ground segmentation that overcomes the presence of gross outliers within the feature matching results~\cite{lim2022single, lim2023quatro++},
and (iii)~hierarchical multi-session SLAM that not only uses our proposed GNC-based registration but also employs a GNC solver to be robust against outlier loop candidates.
Finally, I propose (iv)~instance-aware static map building that can handle the presence of moving objects in the environment~\cite{lim2023erasor2} based on the observation that most moving objects in urban environments are inevitably in contact with the ground.


%
%
%

\section{Towards Robust Perception and Mapping}

\subsection{Noise-Robust Ground Segmentation as Preprocessing}

\begin{figure}[t!]
	\centering
	\captionsetup{font=footnotesize}
	\begin{subfigure}[b]{0.20\textwidth}
		\includegraphics[width=1.0\textwidth]{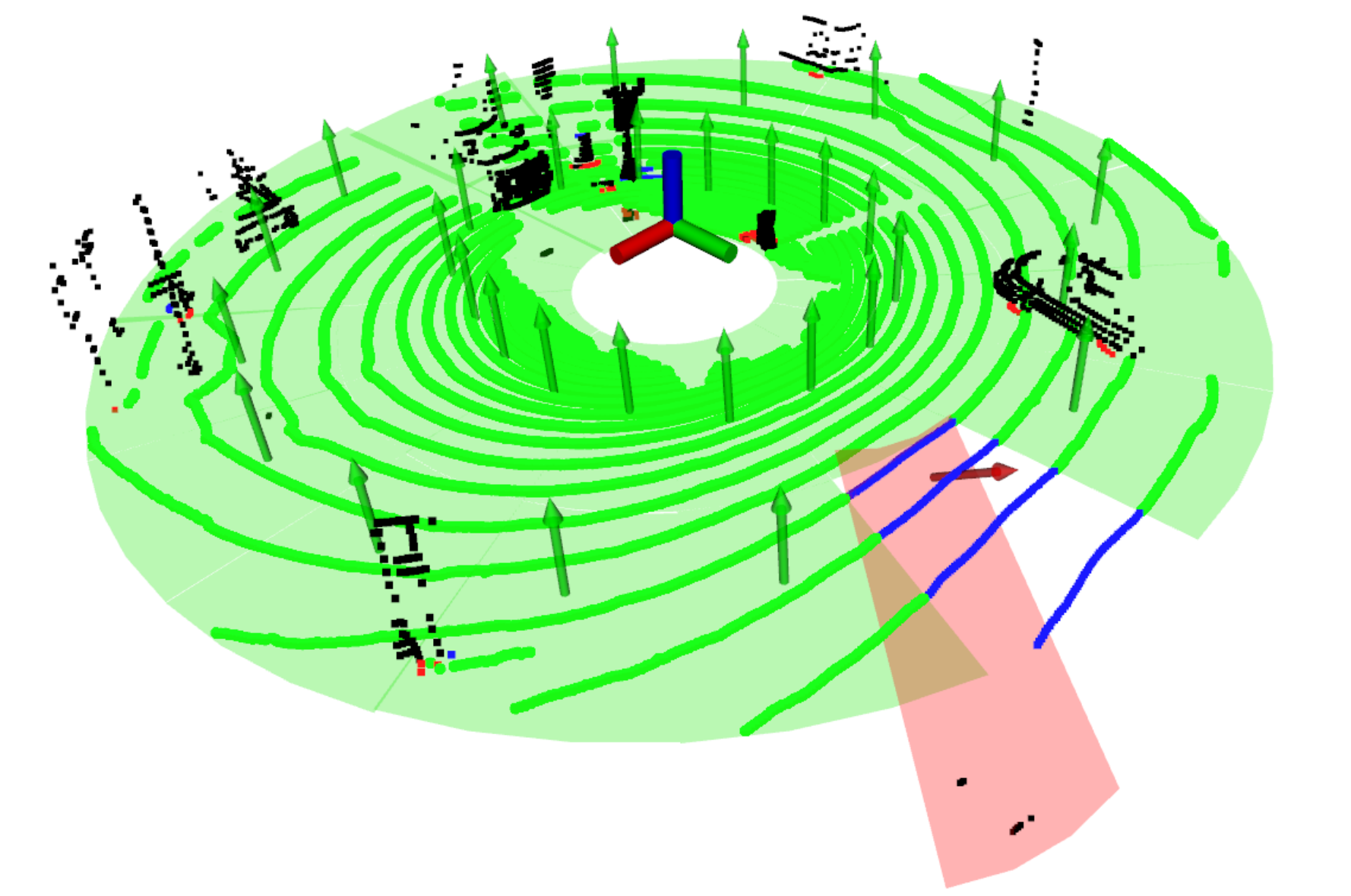}
		\caption{Patchwork++~\cite{lee2022patchworkpp}}
	\end{subfigure}
	\begin{subfigure}[b]{0.20\textwidth}
		\includegraphics[width=1.0\textwidth]{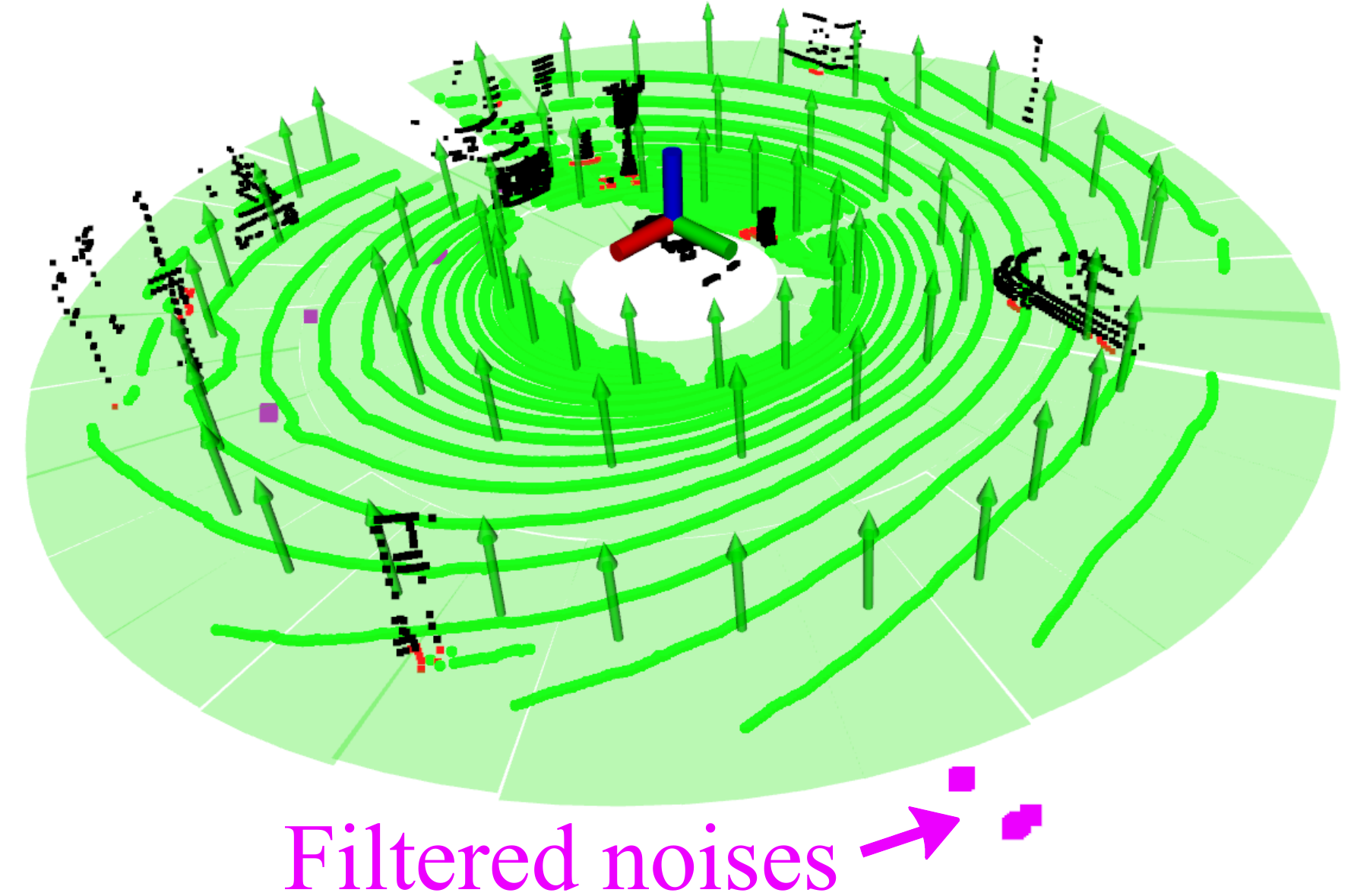}
		\caption{Proposed}
	\end{subfigure}
	\caption{(a)-(b) Performance comparison between Patchwork++~\cite{lee2022patchworkpp} and our proposed approach, which rejects the reflected noises and thus successfully filters ground points. Light green and red bins represent that normal vectors~(arrows) of the estimated planes that are sufficiently upright and too tilted, respectively. Blue points denote wrongly rejected ground points, i.e.,~false negatives~(best viewed in color).} 
	\label{fig:effect_of_outliers}
\end{figure}

\begin{figure}[b!]
	\captionsetup{font=footnotesize}
	\centering
	\begin{subfigure}[b]{0.145\textwidth}
		\includegraphics[width=1\textwidth, trim={0cm 0cm 0cm 14cm}, clip]{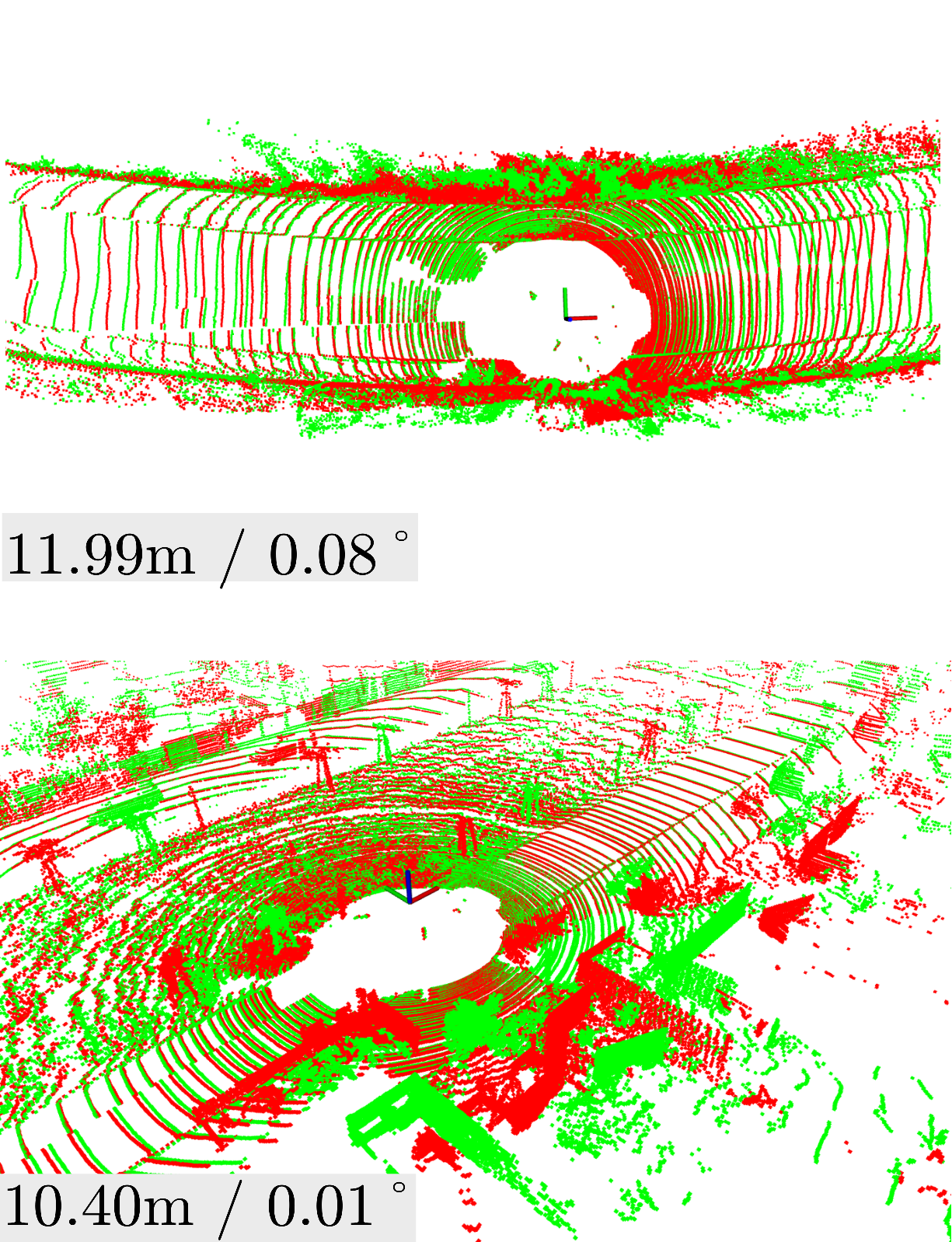}
		\caption{\centering Source and target}
	\end{subfigure}
	\begin{subfigure}[b]{0.145\textwidth}
		\includegraphics[width=1\textwidth, trim={0cm 0cm 0cm 14cm}, clip]{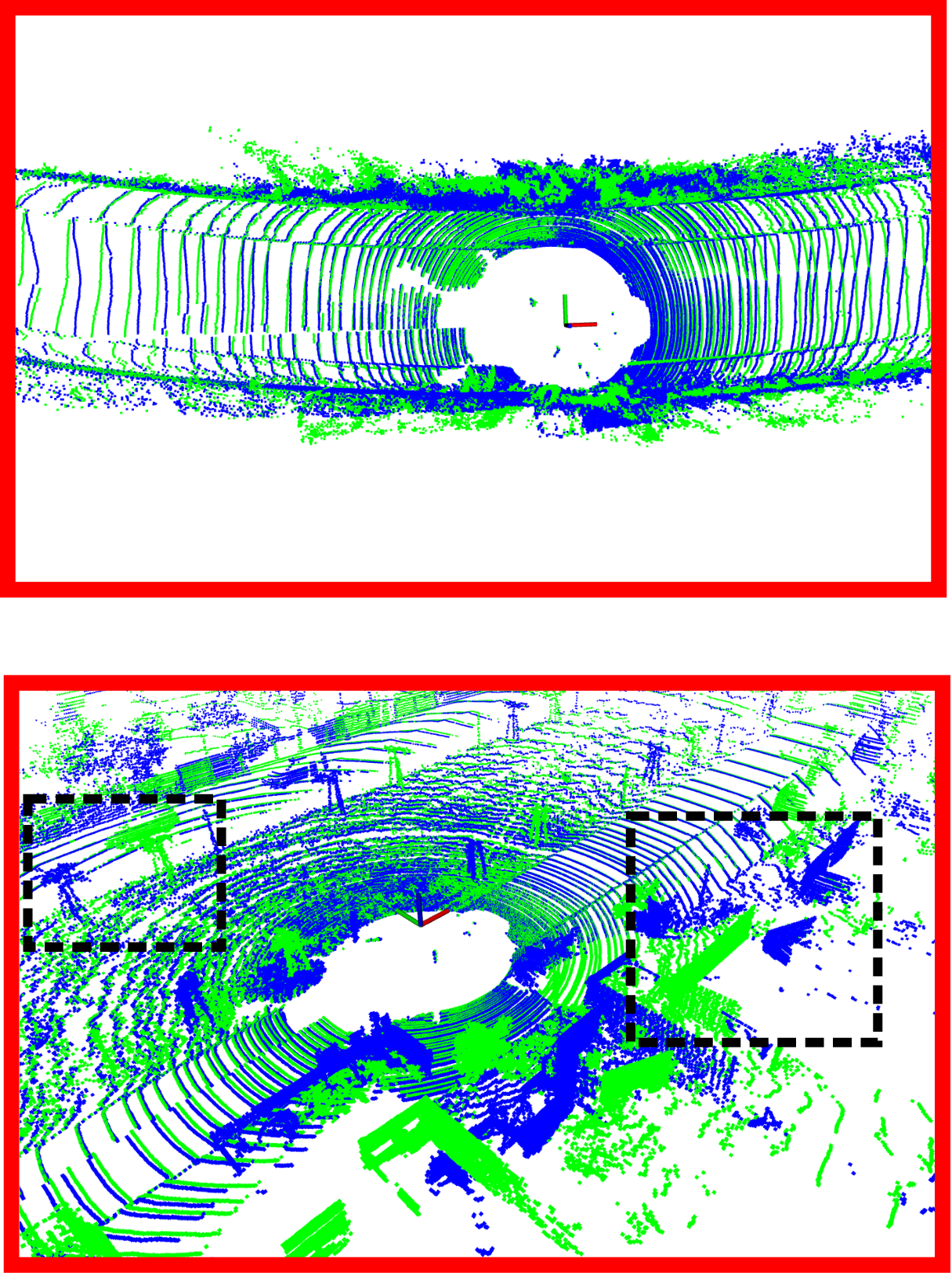}
		\caption{\centering W/o GS~\cite{lim2022single}}
	\end{subfigure}
	\begin{subfigure}[b]{0.145\textwidth}
		\includegraphics[width=1\textwidth, trim={0cm 0cm 0cm 14cm}, clip]{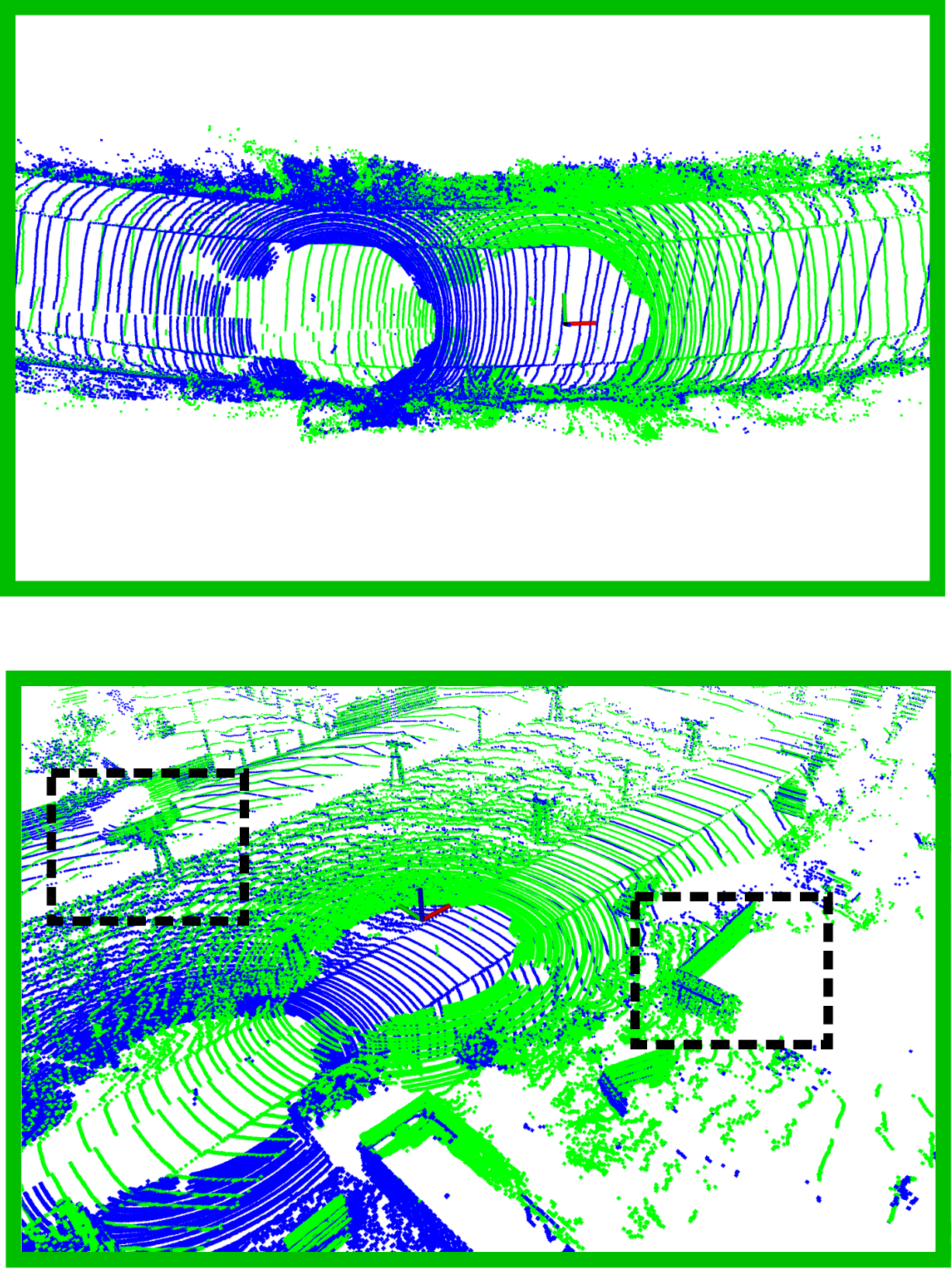}
		\caption{\centering W/ GS~\cite{lim2023quatro++}}
	\end{subfigure}
	\caption{(a) Example of source (red) and target clouds (green) in the KITTI dataset, where the left-bottom text represents the pose discrepancy in each scene. (b)-(c) Registration results before~(W/o GS) and after the application of ground segmentation~(W/ GS), where the blue points denote the warped source cloud by the estimated pose. Our proposed approach~\cite{lim2023quatro++} successfully estimates the relative pose even though the pose discrepancy between viewpoints of source and target clouds is far distant.
	Black dashed boxes highlight the non-ground objects that have to be tightly aligned~(note that, in~(c), these objects are well-aligned). The solid red and green boxes indicate that algorithms failed and succeeded, respectively~(best viewed in color).}
	\label{fig:comp_btw_quatro_and_quatropp}
\end{figure}

First, we propose a fast and robust region-wise ground segmentation~\cite{lee2022patchworkpp, lim2021ral-patch} to be employed as a preprocessing step for instance segmentation~\cite{oh2022travel} or robust registration~\cite{lim2023quatro++}.
In recent years, while ground segmentation methods with high precision and recall in complex urban environments have been proposed~\cite{borges2022survey, mehrabi2021gaussian, proudman2021ecmr, shan2018lego, xue2023jfr},
it has been observed that these approaches sometimes fail due to the noisy points close to the bottom of the ground~(\figref{fig:effect_of_outliers}(a)).
Thus, to address this issue, we filter the noisy points by reusing previously estimated ground planes of neighboring bins when estimating the ground plane of the target bin in a cascaded manner.
Our ground segmentation operates over 100\;Hz for a point cloud acquired by a 64-channel LiDAR sensor while being robust against these noisy points~(\figref{fig:effect_of_outliers}(b));
thus, it can be easily employed as a preprocessing step with little computational burden on the subsequent algorithms.

\begin{figure*}[t!]
	\centering
	\captionsetup{font=footnotesize}
	\begin{subfigure}[b]{0.98\textwidth}
		\includegraphics[width=1.0\textwidth]{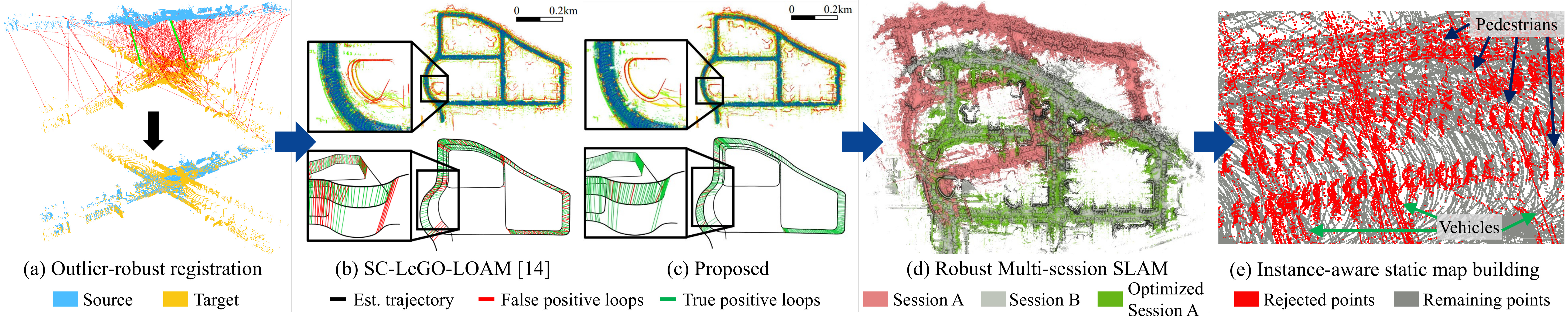}
	\end{subfigure}
	\caption{Our proposed long-term mapping pipeline. (a)~Registration result of our outlier-robust registration framework when estimating the relative pose between two distant viewpoints of point clouds. Our approach robustly estimates the relative pose despite gross outliers~(red lines). (b)-(c) Before and after the application of our proposed registration in the loop closing module of SC-LeGO-LOAM~\cite{kim2018scancontext}. Our proposed registration significantly reduces the number of false positive loops, enhancing the mapping result~(see the upper zoomed boxes). (d)~Our multi-session SLAM results by taking two trajectories from our single-session SLAM as inputs. Note that we leverage our outlier-robust registration to initially align the data via map-to-map registration. (e)~Example of results by the proposed static map building approach. Because the points from moving objects are rejected at an instance-level, traces of pedestrians and vehicles are successfully filtered out. Note that all the approaches are designed to be learning-free and robust; thus, the proposed approaches can be easily employed in various robot platforms and sensor configurations.}
	\label{fig:overview}
\end{figure*}


\subsection{Outlier-Robust Registration for Loop Closing in Single-Session SLAM and Initial Alignment of Multi-Session SLAM}

LiDAR point clouds have sparse characteristics, which means that the regions closer to the sensor origin are denser, and those further away become dramatically sparser~\cite{lim2021ral-patch}.
The issue leads to spurious correspondences when the viewpoint difference between two point clouds becomes distant, which may result in the failure of the registration process~(\figref{fig:comp_btw_quatro_and_quatropp}(b)).
That is, as the viewpoint difference between two point clouds becomes distant, the geometrically distinguishable non-ground points are partially overlapped, and moreover, those overlapped areas are too sparse to establish correct correspondences.
Meanwhile, the ground points near the sensor origin are dominant and geometrically indistinguishable; thus, the feature descriptors from them become ambiguous, resulting in gross outliers~\cite{lim2023quatro++}.

To tackle this problem in a simple yet effective manner,
ground segmentation is exploited as an outlier pruning step and then a GNC~\cite{yang2020teaser}-based robust global registration is applied, which estimates pose while rejecting the outlier measurements simultaneously~\citep{black1996unification,zhou2016fast}, overcoming up to 70-80\% of outliers and providing faster speed compared with previous works~\cite{hartley2009global, pan2019multibnb}.
Consequently, we have demonstrated that our proposed method shows a high success rate when distant source and target clouds are given,
particularly, our proposed approach enhances the PGO results compared with the baseline approach~\cite{kim2018scancontext}~(see Figs.~\ref{fig:overview}(b) and (c)).

Beyond scan-to-scan registration, the proposed method is also applicable to map-to-map registration for initial alignment, owing to its learning-free nature and scalability.
Therefore, our approach enhances the performance of multi-session SLAM by providing a precise initial alignment~(\figref{fig:overview}(d)).

During PGO in both single-session and multi-session SLAM procedures, a GNC solver is also employed~\cite{abate2024kimera2}. By doing so, we can finally achieve consistent and accurate mapping results by suppressing the impact of false positives inter- and intra-loops.

\subsection{Instance-Aware Static Map Building}

Finally, static map building is performed using the scan data and the optimized poses.
As an extension of our previous work~\cite{lim2021ral}, which only detects the traces of moving objects and removes them without the instance-level consideration, we propose instance-level dynamic points removal~\cite{lim2023erasor2}, designed to precisely reject points from moving objects while preserving as many static points as possible.
This approach, based on the observation that dynamic objects move over time and temporarily occupy space, allows us to detect traces of dynamic objects by comparing geometrical discrepancies, i.e.,~the min-max height difference within a grid, between the $t$-th scan and the map cloud.
As a result, we can construct a static map that is directly usable for localization and path planning, as shown in \figref{fig:overview}(e).

\section{Future Research Direction:\\ Semantics-Aware Long-Term Mapping}

\begin{figure}[t!]
	\centering
	\captionsetup{font=footnotesize}
	\begin{subfigure}[b]{0.19\textwidth}
		\includegraphics[width=1.0\textwidth]{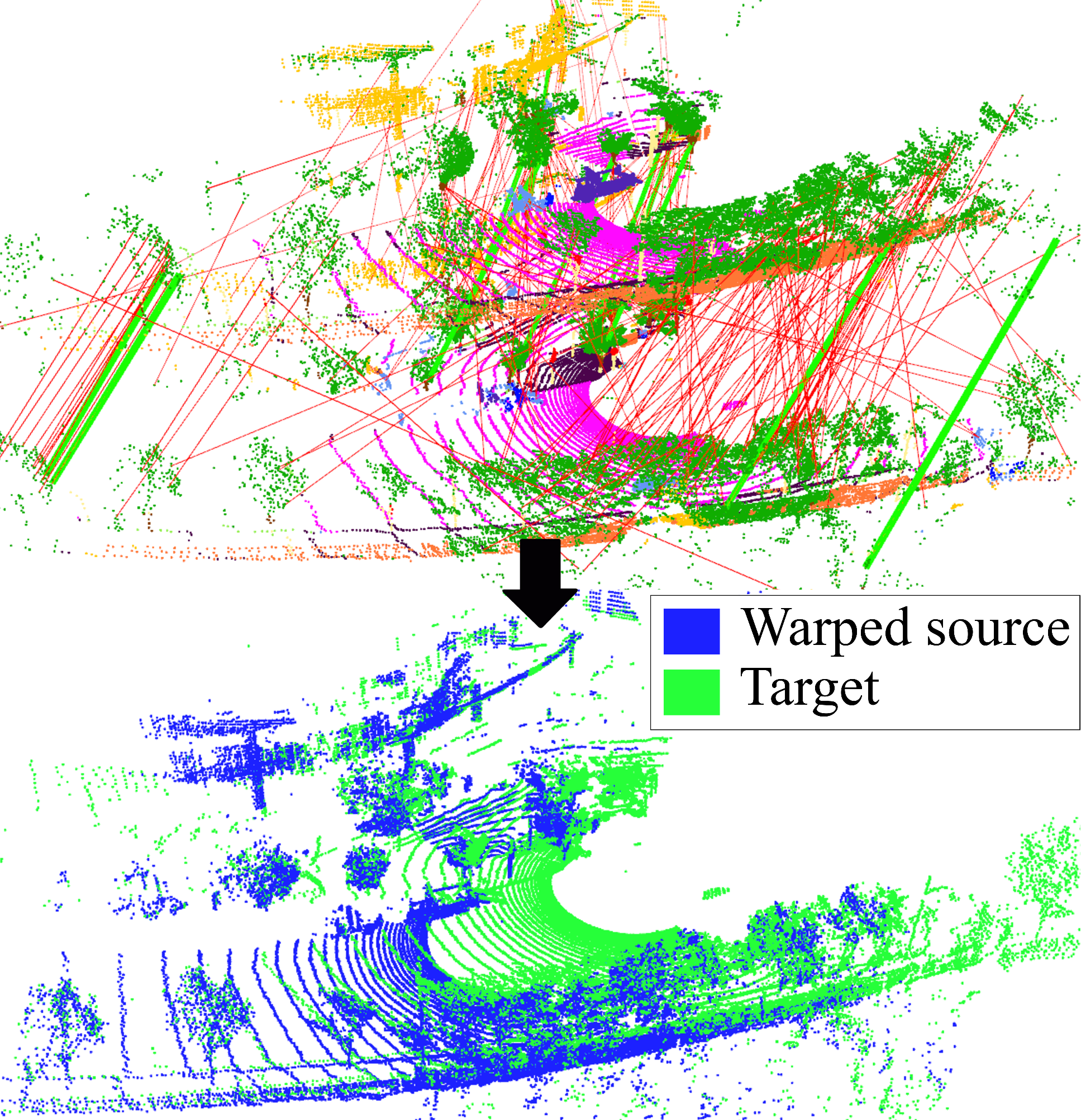}
		\caption{}
	\end{subfigure}
	\begin{subfigure}[b]{0.27\textwidth}
		\includegraphics[width=1.0\textwidth]{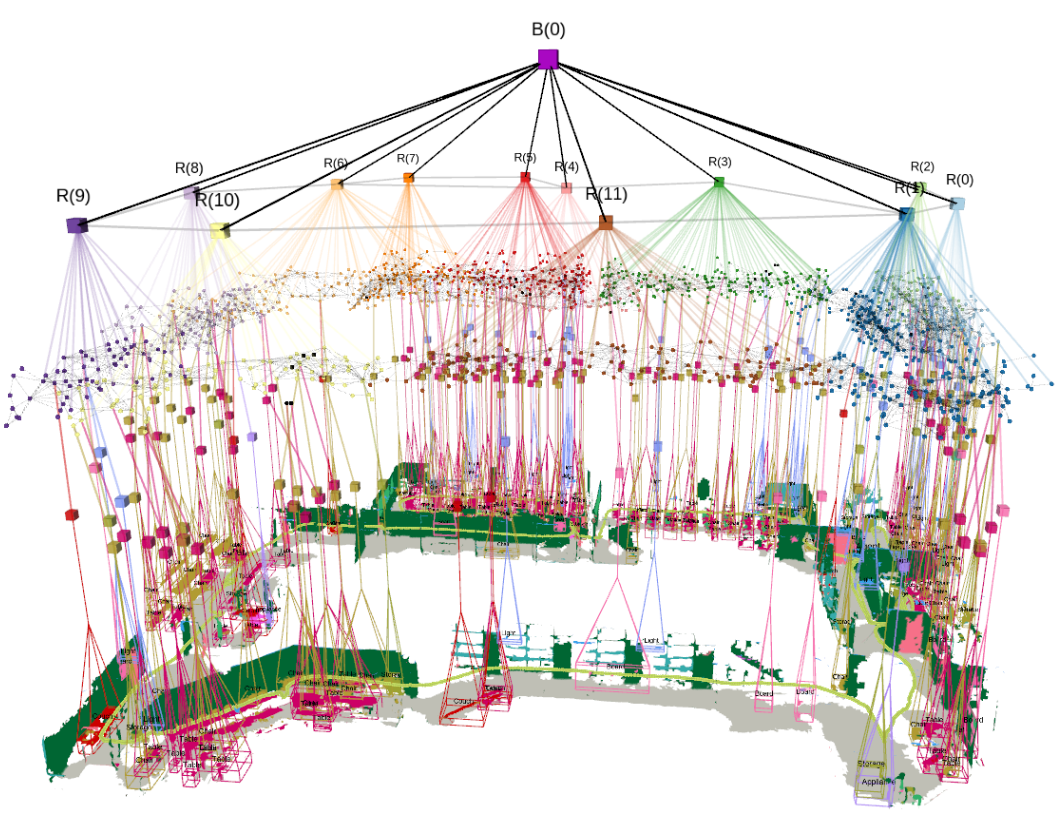}
		\caption{}
	\end{subfigure}
	\caption{Ongoing research topics: (a)~semantics-aware registration output in the MulRan dataset~\cite{kim2020mulran} using the semantic segmentation network trained in the SemanticKITTI dataset~\cite{behley2019iccv} and (b)~scene graph-based mapping result in our campus with zero-shot predictions~\cite{sun2019high}.}
	\label{fig:future_works}
\end{figure}

So far my focus has been on the conventional geometric estimation approaches,
yet I believe integrating semantic information into the geometrical perception is crucial for developing more intelligent and adaptive robotic systems to significantly enhance the robustness and scene understanding capabilities.
I especially advocate for approaches that can robustly perform estimation despite the performance degradation of semantic segmentation in zero-shot situations,~i.e.,~when using the network in unmodeled scenes.

To this end, I embark on two directions of research to integrate semantic information: (i)~semantics-aware registration~(\figref{fig:future_works}(a)) and (ii)~scene graph-based mapping framework~(\figref{fig:future_works}(b)) using zero-shot predictions from the semantic segmentation network~\cite{sun2019high, tang2020searching}.
Consequently, these semantics-aware mapping frameworks would enable robots to navigate and interact with their surroundings in a more intelligent and context-aware manner, significantly boosting their autonomy and effectiveness in performing a wide array of tasks.


\newpage

\bibliographystyle{plainnat}

\bibliography{glorified.bib}

\end{document}